# Interpretable Motion Planner for Urban Driving via Hierarchical Imitation Learning

Bikun Wang, Zhipeng Wang, Chenhao Zhu, Zhiqiang Zhang, Zhichen Wang, Penghong Lin, Jingchu Liu and Qian Zhang.

Horizon Robotics

*Abstract*—Learning-based approaches have achieved remarkable performance in the domain of autonomous driving. Leveraging the impressive ability of neural networks and large amounts of human driving data, complex patterns and rules of driving behavior can be encoded as a model to benefit the autonomous driving system. Besides, an increasing number of data-driven works have been studied in the decision-making and motion planning module. However, the reliability and the stability of the neural network is still full of uncertainty. In this paper, we introduce a hierarchical planning architecture including a high-level grid-based behavior planner and a low-level trajectory planner, which is highly interpretable and controllable. As the high-level planner is responsible for finding a consistent route, the low-level planner generates a feasible trajectory. We evaluate our method both in closed-loop simulation and real world driving, and demonstrate the neural network planner has outstanding performance in complex urban autonomous driving scenarios.

## I. INTRODUCTION

Autonomous driving (AD) in complex urban environments is full of challenges, especially the decision-making and motion planning module in modern autonomy stacks. The mainstream solutions of the decision-making and motion planning module remain traditional rule-based or optimization-based methods, which are difficult to handle with the large number of driving rules and computing resources. Learning-based methods develop rapidly and imitation learning approaches seem the most likely promising way to solve the bottleneck in decision-making and motion planning modules in the short-term.

The main idea behind imitation learning is to use expert demonstrations to learn either a cost function or a direct policy. Taking advantage of abundant supervised learning research achievement, popular imitation learning approaches such as behavior cloning already have many applications due to stable and controllable training process. Besides, large amounts of human driving data as experts' demonstrations promote the development of imitation learning in AD task.

The scope of a qualified data-driven planner in AD system is ambiguous due to complex state-action space and plenty of sensor signals. Most imitation learning work in AD system is to plan with a specific route or an explicit goal. In practice, generating a lane-level route or a pixel-level goal is also a challenging and costly problem. Moreover, another challenge of the learning-based method is using the neural network as a black box lack of security and interpretability, especially end-to-end learning studies.

In this paper, we propose a value-based trajectory transformer VTT, containing a high-level path planner and a low-level executable trajectory planner based on transformer architecture. Integrating decision-making and motion planning as a unified architecture, VTT generates an executable long time horizon trajectory with interpretable spatial and temporal cost maps. The model inputs are middle-level bird's-eye view (BEV) rasterized representation including ego centric road map rasterization and obstacle agents' occupancy. Besides, maneuver waypoints or goal candidates from navigation system are also required as condition input. Both inputs are in rasterized format as in Fig1. Therefore, the planner will generate a reference path and trajectory conditioned on the goal candidate regions.

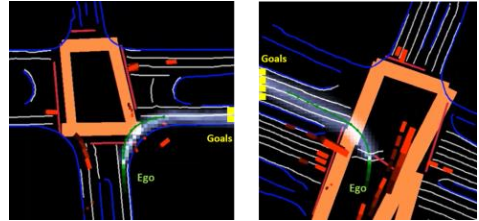

Fig1. Input and output representation of the system. The BEV inputs are from camera perception results. The white transparent render is the behavior distribution as reference path and the green line is the final trajectory in 6s.

Summarize the main contributions of this paper：

- A novel hierarchical transformer-based imitation learning architecture that learns a safe high-level path-selection policy and a low-level trajectory generator policy with simple low-definition arbitrary goal candidates.

- Jointly generating both executable trajectory and cost map, our model is highly interpretable and convenient for rule-based intervention, which makes the system more secure and controllable.

- We demonstrate the effectiveness of our approach in real-world driving in the urban environment of downtown.

## II. RELATED WORK

*Imitation learning* is widely used in robotics and machine learning literature and there are more and more research applied in AD task. The main stream of the method is learning a direct policy generating a trajectory or low-level control command [4][17][24][26][29][30][36], and most of the works use mid-level BEV representation as the input state space. Some other works learn pixel-wise level cost maps and generate the trajectory by solving the cost map function [19][22][37]. Besides, end-to-end research using raw camera pictures as inputs are also becoming popular recently [34][38], which is more challenging due to complex high-level semantic representation learning. Moreover, most of the above tasks are evaluated in simulator and few of them are tested in real-world driving.

***Reinforcement learning*** is another potential application in autonomous driving to optimize the imitation model with proper reward design. It will alleviate the imitation data shift phenomenon and leverage the model learning out of data distribution. Combining imitation learning and reinforcement learning is trend in autonomous driving studies [39][40][41]. However, the quality and realism of the simulator seems to be the bottleneck of these applications.

***Trajectory Forecasting*** Both focus on the executable vehicle trajectory, trajectory forecasting task using deep learning method develop more rapidly [5][6][8][9][10][11][12][13][14][15][16]. The difference between planning task and ego forecast task is that the planning task is an action condition task with navigation and considers more about the trajectory feasibility. Some works predict the agents' trajectory in a planning way using imitation learning or inverse reinforcement learning method [5][16]. Our work is inspired by these planning-prediction research, especially the P2T method as in [5]. We will introduce the comparison between our work and the original P2T work in the section afterwards.

***Transformer*** obtain impressive results in deep learning areas. The transformer architecture has more global and complex encoding ability and some transformer-based planning tasks get outstanding performance [42][43][44]. Our work is also based on transformer encoder and proves that transformer encoding achieve better performance.

## III. BACKGROUND

### A. Markov Decision Process

The problem of behavior planning in AD task can be simplified to the occupancy grid path planning problem based on a markov decision process (MDP), $\mathcal{M} = \{\mathcal{S}, \mathcal{A}, \mathcal{T}, r, \gamma\}$ with a time horizon N. The state S represents the spatial 2-D cells space on the BEV inputs. The action space is grid moving actions function in adjacent grids including {up, down, left, right, up-left, up-right, down-left, down-right, end}. With a deterministic transition function $\mathcal{T}$, $r$ is the reward and $\gamma$ is the discount factor. Given the goal condition and initial state, the learning policy $\pi$ is designed to find the maximum value which is the expected discounted sum of rewards over a fixed time horizon as in formula (1).

$$V^\pi(s) = \mathbb{E}^\pi \left[ \sum_{t=0}^{T} \gamma^t r(s_{t'}\, a_t) \right] \tag{1}$$

### B. Value Iteration Network

Since the convolutional neural networks using kernels traverse the spatial feature, the MDP can be expressed as convolutional calculations as in [1]. Using a convolutional 3x3 kernel weights as the deterministic transition of the 9 actions, the expected value is calculated iteratively from the goal state to the start state, which is typically an approximation of dynamic programming application. Then the spatial reward map and the gird policy will be learned from expert demonstrations as in formula (2). As an imitation learning policy, the state-action pair from offline expert datasets $D$ has the maximum likelihood distribution.

$$\pi_\theta(a|s) = \underset{\pi_\theta}{argmax} \sum_{s,a \in D} \pi(a|s) \tag{2}$$

The whole value iteration network (VIN) block in [1] utilizes a fully differentiable network to solve the grid path finding problem. With stable supervised training process, we leverage the VIN block in our model to learn a reference path distribution as described in section below.

## IV. METHODOLOGY

In this section we describe VTT including input representation, high-level route selection policy based on MDP and low-level continuous trajectory policy. The whole picture of the model is shown as in Fig.2. Finally, we describe the loss design and the training strategy.

### A. Model Architecture

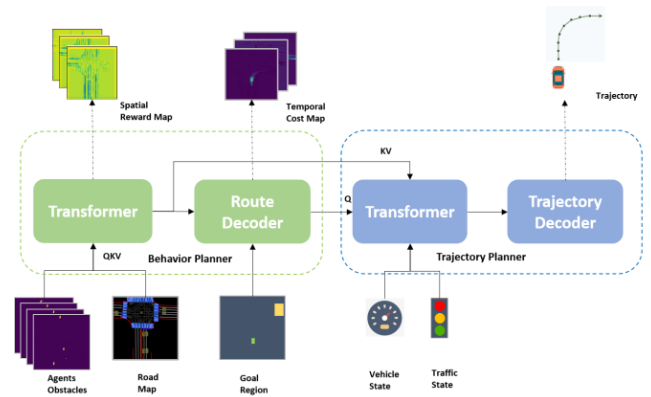

Fig2. Schematic diagram of the proposed value trajectory transformer (VTT) model architecture.

Considering interpretability and practicability, our work focuses on mid-to-mid tasks with BEV inputs and generates both trajectory and cost maps. The rasterized inputs could either come from HD-map with a global location coordinate, or from the upstream perception module results which is decoded from the raw camera. With the perception results as inputs, our method is closer to an end to end study. In order to reduce the inputs' interference and focus on the neural planner study, we use HD-map results as the inputs in the experiments afterwards.

*Input representation:*

- Rasterized BEV map: current ego-centric images containing static road environment and dynamic obstacles occupancy.

- Goal Regions: achieved from navigation system, low-definition pixel blocks on a binary image which indicate the probable driving direction on the road map.

- Dynamic states: traffic lights state and ego speed.

*Output representation:*

- Cost maps: heat-map of the spatial and temporal experts' intent probability distribution, which is generated from the high-level planner. These cost maps are easily combined with a rule-based planner to keep the safety of the whole system.
- Trajectory: a long time horizon trajectory is generated in 6 seconds with a 2Hz sampling frequency. The number of waypoints is the same as the dynamic cost map channels.

### B. High-level Behavior Planner

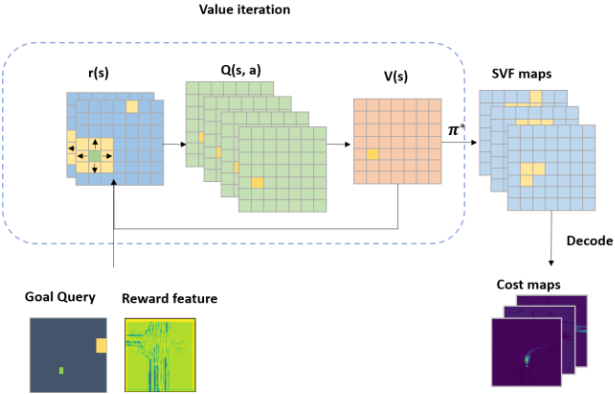

Fig3. Simple schematic diagram of the route decoder.

In the high-level module, the VIN block is used for generating the spatial value map by calculating the reward map iteratively as in Fig3. Firstly, the reward feature map are generated from the upper transformer encoding block with the BEV inputs. With strong spatial modeling ability, the cross-attention mechanism learns the reward function better than CNN architecture. Concatenating the rasterized goal region map, the reward feature maps are sent to the VIN block. Secondly, with the approximate value iteration, we get a goal conditioned grid transition policy which is non-stationary at different time horizon k. The policy $\pi$ indicates the action probability at any location of the grid map. Therefore, an arbitrary starting point can propagate through the policy in a certain horizon. The value iteration process is defined as in formula (3).

$$\begin{cases} Q^k(s',a) = r_\theta(s) + V^k(s') \\ V^{k-1}(s) = \log \sum_a exp^{Q^k(s,a)} \quad s' = T(s,a) \\ \pi_\theta^k = exp^{Q^k(s,a) - v^k(s)} \end{cases} \quad (3)$$

With policy propagation, the route path probability in the grid map can be generated by calculating state visit frequency (SVF) $P^k(s)$ step by step. Starting from the initial ego state, the states' distribution in k steps is represented in k steps SVFs which are spatially related by single step grid transition as in formula (4). Finally, we add a simple decoder make the SVFs into temporal cost maps and the channel of the cost maps is same as the final trajectory horizon correspondingly. The simple decoder contains some basic convolutional layers which is responsible for combining the grid-based spatial SVF maps into time related attention maps.

$$P^{k+1}(s') = \sum_{s,a} \pi_\theta^k(a|s) P^k(s) \quad (4)$$

The VIN block guarantee the consistency of the SVF probability distribution. With deterministic goals, the corresponding path could always be found and the whole VIN block do not have any learnable weights. The high-level behavior planner provide a lane-level solved route and future temporal ego occupancy. Given approximate location of future ego, these cost maps are not only interpretable output and also the query of the low-level planner transformer encoding input.

### C. Low-level Trajectory Planner

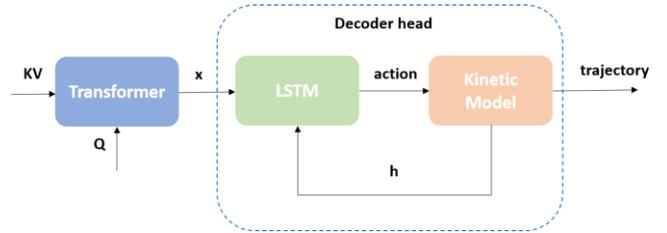

Fig4. RNN-based trajectory head. KV(key-value) is the high-level feature and the Q(query) is the learned cost maps

Considering cost maps as query of the low-level transformer input, the other inputs are from the high-level transformer output feature embedding fusing dynamic states input. In the low-level module, we employ LSTM block as the trajectory head which is a popular recurrent neural network (RNN) block. The LSTM encoder realize the hidden state transition and produces direct control signal acceleration and curvature. Finally, a simple kinetic bicycle model will calculate the corresponding specific waypoints.

$$\begin{cases} x_{t+1} = x_t + v_t cos(\theta_t)\Delta t \\ y_{t+1} = y_t + v_t sin(\theta_t)\Delta t \\ \theta_{t+1} = \theta_t + v_t \kappa_t \Delta t \\ v_{t+1} = v_t + a_t \Delta t \end{cases} \quad \kappa_t \approx \frac{tan(\delta_t)}{L} \quad (5)$$

The trajectory head is as in Fig4 and the RNN architecture will make the adjacent waypoints temporal related to each other. Same as [24][35], the kinetic model in formula (5) makes the trajectory more feasible and controllable. Considering the control action outputs as waypoints delta incensement, we can easily apply the control constraints to make the trajectory more executable.

### D. Loss Design

Both the high-level and the low-level policy are supervised by expert demonstrations, and The total loss consists of four main components as below, including high planner loss, low planner loss, cost loss and auxiliary loss, and $\lambda$ is a tunable parameter.

$$\mathcal{L} = \lambda_h \mathcal{L}_{high} + \lambda_l \mathcal{L}_{low} + \lambda_c \mathcal{L}_{cost} + \lambda_a \mathcal{L}_{aux} \quad (6)$$

The high loss is the normal behavior clone loss, which aims to reduce the distribution gap of the learning weights and the experts as in formula (7). We use L1 loss to mimic the experts' trajectories, and the output states contain the waypoint coordinates, yaw angle and ego speed as defined $s_t = \{x_{t'}, y_{t'}, \theta_{t'}, v_t\}$. Since the trajectory is generated from the kinetic model, a L2 regularization penalty is applied to the network output to keep the smoothness of the trajectory, which is actually constraint of the acceleration and the curvature as in formula (8).

$$\mathcal{L}_{high} = \mathbb{E}_{s,a \in \pi_\theta} \log \pi_\theta(a|s) \qquad (7)$$

$$\mathcal{L}_{low} = \sum_{t=1}^{K} ||s_t - s_t^{gt}||_1 + \alpha ||a_t||_2 + \beta ||\kappa_t||_2 \qquad (8)$$

After the policy propagation and route decoding, the dynamic costs are supervised using pixel-wise focal loss as $\mathcal{H}$ which represent the probability distribution in a specific time step. The time horizon is same as the trajectory output. The ground truth of the cost map is the expert occupancy grid in different time steps.

$$\mathcal{L}_{cost} = \sum_{t=1}^{K} \mathcal{H}\left(C_t(x, y), C_t^{gt}(x, y)\right) \qquad (9)$$

$$\mathcal{L}_{aux} = \sum_{t=1}^{K} \sum_{i=1}^{N} \mathcal{G}(s_t)_i \mathcal{T}_i^{raster} \qquad (10)$$

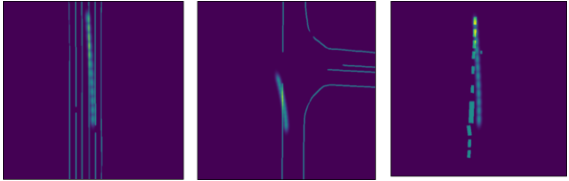

Fig5.Schematic diagram of differentiable rasterized auxiliary loss. The losses in the pictures aim to penalize the abnormal behaviors.

In order to decrease the occurrence of dangerous behaviors, we calculate potential off-road and collision probability using the rasterized trajectory mask with the environment mask, which is the auxiliary loss same as in [20][24]. Every trajectory waypoint $s_t$ is rasterized into an occupancy grid through ellipse Gaussian kernels with the vehicle length and width, where $\mathcal{G}(s_t) = \mathcal{N}(\mu, \Sigma)$ is the rasterization function to convert trajectory into the vehicle box-like shape occupancy as in Fig5. $\mathcal{T}_i^{raster}$ is the given binary masks including the static road topology and dynamic obstacles. The calculation of the rasterized auxiliary loss is defined in formula (10).

### E. Data Augmentation

Proper data augmentation makes policy learning from the data out of the experts' demonstration distribution. We disturb the initial start waypoint of the trajectory with a random displacement and heading angle similar to [4]. Then update the BEV inputs with the perturbation error correspondingly. Since the perturbation start point and the fixed endpoint is known, we use a lightweight optimizer to calculate the new trajectory ground truth with motion constraints which keep the new trajectory is the feasible action to recover from the perturbation error. Since the Dagger[2] methods alleviate the distribution shift problem of behavior clone method, we train the network in a close-loop way. After every training epoch, the performance of the current learned policy in the experts datasets will generate new synthesized data samples with the calibrated trajectory ground truth which is also generated by the optimizer as mentioned above. The key difference from the perturbation method is the noise coming from the outside randomly or current policy itself. The optimizer plays an important role as a queryable expert which help fix the uncorrected trajectory.

## V. EXPERIMENTS

### A. open-loop evaluation

We evaluate our method on real-world driving datasets collected in the urban environment called TDT datasets. A variety of urban driving scenarios data was collected including multi-lanes and intersections for more than 50 hours. The model inputs fields of view of including the road map and the 2s dynamic agents occupancy rasterization are 102.4m x 102.4m with 0.2m/pixel resolution, which is 512x512 corresponding to the image size. The position of the ego car is fixed at (362，256) of the image with 72.4m in front and 30m in back and both 51.2m to the left and right. The output trajectory is 2Hz with total 6s planning horizon, and the time step between every adjacent waypoint is 0.5s. Although the history information of the ego vehicle will improve the open-loop results especially the expert similarity, we do not use any ego history information which will lead to causal confusion as mentioned in [3]. The causal confusion detail of imitation learning is not discussed in this paper.

The open-loop evaluation mainly reveals the network learning ability of the similarity to the experts' distribution. We use L1 distance between the predicted path from high-level policy propagation and the expert binary distribution to metric the high-level planner's performance (See svf diff in Table I) . We use L2 distance between the learning trajectory and the expert trajectory to represent the expert similarity, which is popular used in prediction task and known as ADE(average displacement error). Also, we calculate the future ego occupancy along with the predicted trajectory to check the intersection with the future obstacles and the road edge to represent the potential crash rate and the off-road rate. Every way point is calculated separately in the specific time horizon.

TABLE I.          OPENLOOP TEST RESULTS

| Base line | SVF Diff (%) | ADE(m) | | Crash rate (%) | | Off-Road rate (%) | |
|---|---|---|---|---|---|---|---|
| | | ave | final | ave | final | ave | final |
| P2T | 0.80 | 1.48 | 3.58 | 0.71 | 1.80 | 0.12 | 0.2 |
| VTN | 0.85 | 1.47 | 3.56 | 0.70 | 1.72 | 0.09 | 0.22 |
| **VTT** | **0.73** | **1.40** | **3,35** | **0.63** | **1.68** | **0.05** | **0.22** |

From the table, the proposed approach VTT achieves the best performance from all sides. The P2T baseline is our first baseline using the method [5] containing a reward model and a trajectory generator. However inputs of the low-level policy are sampled and cropped ROI features by the high-level policy instead of learning costs, which could not trained end to end. We implement the goal-conditioned P2T method in our TDT datasets using the open source code. The VTN baseline is the same as VTT with only CNN encoding rather than transformer.

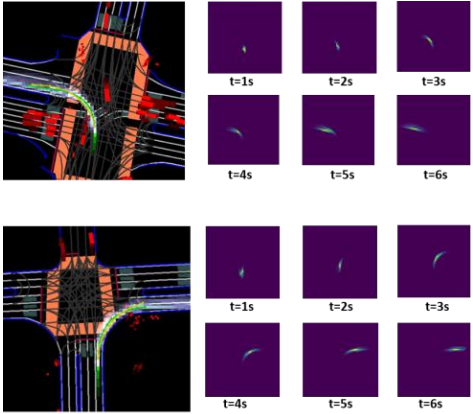

Fig6 visualization of planning outputs in different time horizon

Interpretability is very significant in the driving system and it not only help explain neural network learned distribution but also benefit to the rule-based planning system. The goal conditioned VIN block in high-level planner guarantee that there is always a consistent path could be found to the goal regions. The cost occupancy maps are located inside the consistent path which indicate the most likely position where the experts will pass as in fig5. The white transparent area in the left picture is the learned route path by VIN and the yellow trajectory is the low-level learned trajectory as the green is the ground truth. The cost maps in Fig6 represents the learned occurrence distribution at a specific time step from 1s to 6s. These middle cost map outputs can not only help with bad case explanation study, but also are convenient to add rule-based manual intervention as the goal-keeper of the whole system. For example, these cost maps could be used as the inputs of a simple rule-based trajectory sampler, and help to evaluate the optimized trajectory.

### B. close-loop evaluation

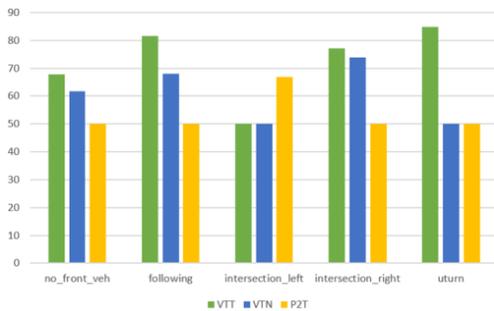

Fig7 close-loop simulation results in typical scenarios

We employee two types of closed-loop simulations to evaluate the planner performance, world-sim and log-sim respectively. In the world-sim, static scenarios are generated inside a preset environment and the agents are rule-based intelligent driver model (IDM). Relatively, scene information in the log-sim are extracted from field tests and replayed with minor modification in the simulation. In this study, log-sim is taken as the major approach to evaluate the planner's close-loop performance under consideration that field test data provides us with more real and complex scenarios. Log data used in the simulation are clipped from field tests with expert trajectories, in which each lasts tens of seconds and various scenarios are covered, such as following, right turn, left turn, U-turn, lane changing etc.

The close-loop evaluation design reference CARLA metric. Leveraging the availability of expert data in the log-sim, the major metrics are route completeness, trajectory similarity and overall performance. Instead of a L1 metric in open-loop, the trajectory similarity is calculated through lock-step Euclidean distance method to evaluate the similarity between two trajectories. We use the final driving score (DS) as the overall close-loop evaluation metric as in formula (11). $RC_i$ is the route completeness which is the ratio of the simulated ego travel path to the experts' route. $P_i = \prod_{j=1}^{n} \lambda_j^{k_{i,j}}$ is the infraction discount in the i-th simulation including collision with obstacles and traffic rule violations. The total number the infraction types is $n$, where $\lambda_j$ is the discount coefficient of infraction type j and $k_{i,j}$ is the counts of the j infraction in the i-th simulation.

$$DS = \frac{1}{N} \sum_{i=1}^{N} RC_i P_i \tag{11}$$

Same as the open-loop evaluation comparison, the closed-loop simulation performance which compares the VTT model, VTN model and P2T model are shown as in Fig.7. Five scenarios are considered, including cruising without front vehicles, following, left turn, right turn and u turn. Every evaluate scenario contains more than 100 clips log data and the final score is the average performance of all the validate scenarios. From the results, we can see that the VTT model achieve better performance in most driving test scenarios.

### C. Real-World Driving

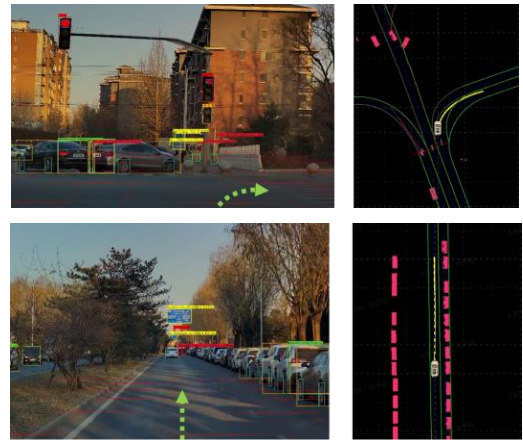

Fig8 example of real world testing scenario

We implement our approach in the real world with Horizon Journey 5 chips as integrated hardware. Our model is quantized into an int8 format with less accuracy loss and achieves 80~100FPS inference speed. Instead of using large amounts of data, we only train our model in our small datasets mentioned before including 50h+ human driving data to validate the model performance. The raw trajectory output from the network is directly translated to control signal through a MPC controller with barely rule constraints. The self-driving system is tested in the downtown environment under the supervision of human safety drivers. Our approach demonstrates positive human-like driving ability facing all kinds of urban driving challenges, including lane-keeping, lane-changing, turning left or right at intersection and U-turn. Especially in the intersection turning scenario, the network trajectory is more comfortable with a lower curvature than the reference drive line, which acts more like a real human driver.

## VI. Conclusion

We proposed an explainable hierarchical neural network motion planner using imitation learning for urban autonomous driving applications. Integrating decision-making and motion planning modules, the high-level policy decodes the goal regions into coarse route occupancy which solve a lane-level path-finding problem, and then the low-level policy generates waypoint trajectory. The hierarchical architecture provides different levels of outputs and guarantees a stable training convergence. We demonstrate our model performance both in simulation and real-world driving and our approach achieves outstanding performance in complex urban driving scenarios. In the future, we will continue to explore the limits of imitation learning by enlarging expert data and focusing on improving generalization performance in extremely challenging driving scenarios. Besides, reinforcement learning is also another way to improve model performance in interaction with static and dynamic agents.